\documentclass[runningheads]{llncs}
\usepackage{float}
\usepackage{graphicx} 
\usepackage{color} 
\usepackage[dvipsnames]{xcolor}
\definecolor{brickorange}{RGB}{193,74,9}
\definecolor{blue2}{RGB}{162,191,254}
\definecolor{red}{RGB}{255,0,0}
\definecolor{green}{RGB}{0,255,0}
\definecolor{green}{RGB}{51,132,51}
\definecolor{pink}{RGB}{249,187,195}
\definecolor{blue}{RGB}{192,247,249}
\definecolor{yellow}{RGB}{250,224,40}
\definecolor{red}{RGB}{222,91,76}
\definecolor{purple}{RGB}{222,156,218}
\definecolor{orange}{RGB}{245,136,24}
\usepackage{caption} 
\usepackage{subfigure} 
\usepackage{wrapfig} 
\usepackage{amsmath,amssymb,marvosym} 
\usepackage{array} 
\usepackage{hyperref}
\newcommand{\etal}{\textit{et al.}}
\usepackage[colorinlistoftodos]{todonotes} 
\usepackage{algorithm} 
\usepackage[noend]{algpseudocode}

\usepackage{pgfplots}
\usepackage{environ}
\makeatletter
\newsavebox{\measure@tikzpicture}
\NewEnviron{scaletikzpicturetowidth}[1]{%
  \def\tikz@width{#1}%
  \begin{lrbox}{\measure@tikzpicture}%
  \BODY
  \end{lrbox}%
  \pgfmathparse{#1/\wd\measure@tikzpicture}%
  \BODY
}
\makeatother
\usepackage{enumitem}
\usepackage{makecell}
\usepackage{url}
\usepackage{wasysym}
\usepackage{multirow}
\usepackage{hyperref}
\usepackage{tikz}
 
\usepackage{hyperref}
\hypersetup{
    colorlinks=true,
    linkcolor=blue,
    filecolor=magenta,      
    urlcolor=cyan,
}
\usepackage{lipsum}
\newcommand{\repeatthanks}{\textsuperscript{\thefootnote}}

\begin{document}
\title{Evaluating the Robustness of Self-Supervised Learning in Medical Imaging}
%
%
\author{Fernando Navarro\thanks{The authors contributed equally to the work. Code \href{https://github.com/ferchonavarro/ss\_medical\_robustness}{https://github.com/ferchonavarro/ss\_medical\_robustness}.}\inst{1,2,3}, Christopher Watanabe\repeatthanks\inst{1},Suprosanna Shit\inst{1,3}, Anjany Sekuboyina\inst{1,3,5} , Jan Peeken\inst{2,4}, Stephanie E. Combs \inst{2}, \and Bjoern H. Menze\inst{1,3,5}}

\titlerunning{Anonymous}  

%

\institute{Department of Informatics And Mathematics, Technische Universit\"at M\"unchen, Germany.\\
\and
Department of Radiation Oncology, Klinikum rechts der Isar, Germany.\\
\and
TranslaTUM - Central Institute for Translational Cancer Research, Munich, Germany.\\
\and
Institute of Radiation Medicine (IRM), Department of Radiation Sciences (DRS), Helmholtz Zentrum M\"unchen, Germany.\\
\and
Department for Quantitative Biomedicine, University of Zurich, Switzerland.\\
\email{fernando.navarro@tum.de}}

\maketitle              
%

%

\begin{abstract}
Self-supervision has demonstrated to be an effective learning strategy when training target tasks on small annotated data-sets. While current research focuses on creating novel pretext tasks to learn meaningful and reusable representations for the target task, these efforts obtain marginal performance gains compared to fully-supervised learning. Meanwhile, little attention has been given to study the robustness of networks trained in a self-supervised manner. In this work, we demonstrate that networks trained via self-supervised learning have superior robustness and generalizability compared to fully-supervised learning in the context of medical imaging. Our experiments on pneumonia detection in X-rays and multi-organ segmentation in CT yield consistent results exposing the hidden benefits of self-supervision for learning robust feature representations.

\end{abstract}

\section{Introduction}

With the advent of deep learning in computer vision, the medical imaging community has rapidly adopted and developed algorithms 
to assist physicians in clinical routine. For instance, deep neural networks trained for classification can aid in the diagnosis of abnormalities such as brain tumors \cite{amin2017distinctive,bahadure2017image}, bone fractures \cite{thian2019convolutional,husseini2019conditioned}, and a range of skin lesions \cite{esteva2017dermatologist,navarro2018webly}. Likewise, deep learning-based segmentation algorithms provide automatic delineation of structures or organs of interest in medical procedures \cite{ibragimov2017segmentation}. Nevertheless, the overall performance of fully-supervised learning highly depends on the size of the annotated data-set, resulting in sub-optimal performance for smaller data-sets. Consequently, state-of-the-art approaches seek to develop novel methods to overcome the problem of learning under a low data regimen.\\

\noindent\textbf{Self-Supervision Literature Review:} Self-supervision has emerged as a potential solution for learning representations without the requirement of labeled data. The vast majority of current work in self-supervision focuses on developing pretext tasks that learn meaningful representations, which can be reused later for the target task with minimum annotations. Self-supervision approaches can be classified into encoder and encoder-decoder architectures. The work introduced in Doersch \etal \cite{doersch2015unsupervised}, Noroozi \etal \cite{norooziUnsupervisedLearningVisual2016}, and Gidaris \etal \cite{gidaris2018unsupervised} are examples of encoder architectures.
These models train the encoder of a convolutional neural network (CNN) to predict the relative position of images patches, solve Jigsaw puzzles, and predict image rotations, respectively. Encoder-decoder architectures train deep neural networks (DNNs) on reconstruction tasks. Chen \etal \cite{CHEN2019101539}, for example,  trained a network on the invented context restoration task, while Zhou \etal \cite{zhouModelsGenesisGeneric2019} explored the idea of generic autodidactic models, which learn to reconstruct missing or perturbed information. In both self-supervised paradigms, the learned image representations are used to initialize either the encoder or encoder-decoder part of a neural network fine-tuned on the target task.\\


\noindent\textbf{Robustness and Generalizability in Self-Supervision:}
Despite efforts to bridge the gap between fully-supervised and self-supervised learning in medical imaging, the latter lag behind the former in standard metrics for both classification and segmentation. Furthermore, when the same training data is used for both self-supervision and fine-tuning, self-supervision yields a marginal impact on performance. The focus on raw performance metrics, however, distracts from other real potential benefits of self-supervision, such as increased robustness and generalizability. This is of significant worth in the medical image context, as it determines the breadth of network applicability over medical data and acquisition centers. Ideally, trained networks should succeed in their respective tasks regardless of the acquisition centers, acquisition devices, image resolution, potential artifacts, or noise present in the data. Toward this goal, we adapt robustness analysis methods from Hendrycks \etal \cite{hendrycks2019robustness,hendrycks2019pretraining,hendrycks2019using}, who demonstrated the robustness of self-supervision in ImageNet classification networks. We aim to ascertain whether self-supervision contributes to the robustness of networks in medical imaging in classification and segmentation tasks.


In the context of medical image analysis, we show that self-supervised gains are masked when looking at the overall performance of a network inferred on clean data, which is similar to the performance of fully-supervised networks. Our results reveal the hidden robustness gains that accompany self-supervision on medical imaging tasks.











\section{Methodology}

\begin{figure}[ht]
    \centering
        \includegraphics[width=1.0\linewidth]{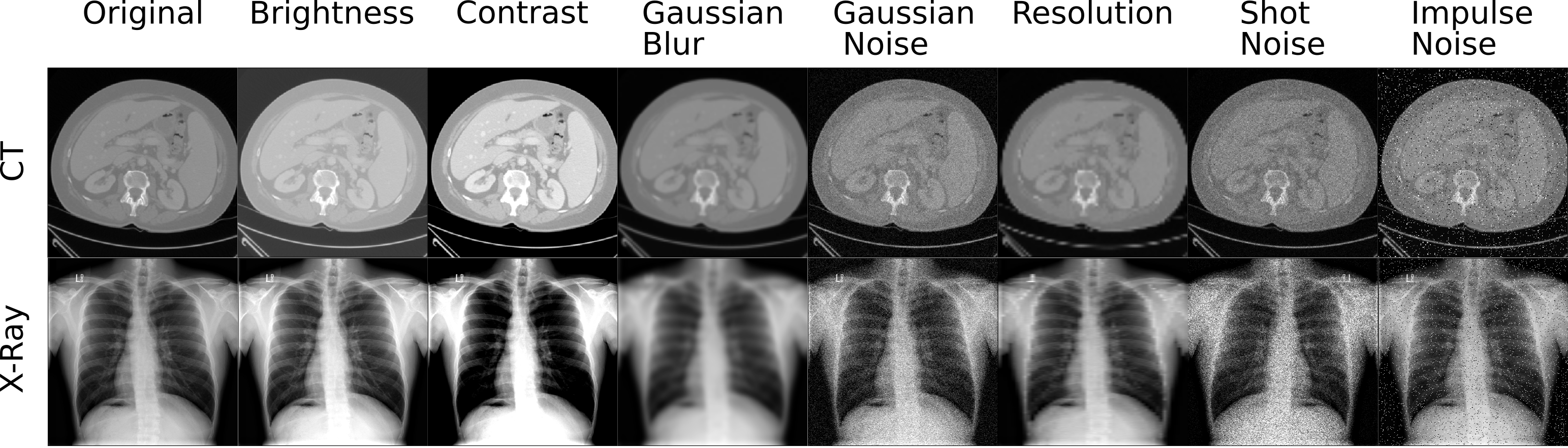}
    \caption{\footnotesize{Examples of image corruptions. The first column shows the original image, while the following columns illustrate different image corruptions. The first row corresponds to a CT image for multi-organ segmentation. The second row shows an example X-ray image for pneumonia detection.}}
    \label{fig:imagecorruption}
\end{figure}

We evaluate and compare the robustness and generalizability of self-supervision and fully-supervised learning for two tasks,  pneumonia detection in X-ray images and multi-organ segmentation in CT images, covering both global and local classification tasks. In the following sections, we describe the followed protocol to compare the robustness and generalizability between self-supervision and fully-supervised learning.\\




\noindent\textbf{Image Corruptions:} Improving robustness to image corruptions was previously studied against adversarial examples \cite{paschali2018generalizability} and common image corruptions \cite{hendrycks2019robustness}. Although adversarial examples are a good basis to benchmark neural networks, they are not common in real-world applications. Hence, we focus on seven image corruptions commonly encountered during acquisition, shown in Fig. \ref{fig:imagecorruption}. In medical imaging, there is always a trade-off between minimizing radiation doses at the cost of image noise and quality. Examples of such noises include Gaussian noise, shot noise and impulse noise, which are included in our simulated image corruptions. Changes in image appearance are simulated by image contrast and brightness, as a surrogate of contrast enhancement and high exposure. Gaussian blur simulates patient movement during acquisition and resolution refers to data coming from different scanner resolutions. 

Formally, let $x$ be a clean image in data-set $\mathcal{D}$. We define $g(\tilde{x}\mid x,c,s): x\rightarrow \tilde{x}$ as a function that transforms sample $x$ into $\tilde{x}$ according to a corruption $c\in C$, where $C$ is the set of corruptions, and $s$ is the severity of the corruption.\\

\noindent\textbf{Robustness:} Let's denote the posterior probabilities on a corrupted sample $\tilde{x}$ of a fully-supervised trained network $f(y\mid \tilde{x}, \theta_{1})$ and a self-supervised trained network  $h(y\mid \tilde{x}, \theta_{2})$ . We declare that $h(y\mid \tilde{x}, \theta_{2})$ is more robust if the learned feature representations are less sensitive to the image corruption function $g(\tilde{x}\mid x,c,s)$ with respect to an evaluation metric $m$.\\

\noindent\textbf{Generalizability:} The generalizability of a trained neural network can be assessed by deploying the network to an unseen data-set. Ideally, a network is able to generalize even when the data comes from a different distribution. In the context of medical imaging, this translates to the capability of a network and its feature representations to be robust when deployed in a data-set coming from a different health center or different scanning device.\\


\begin{figure}[ht]
    \centering
        \includegraphics[width=0.8\linewidth]{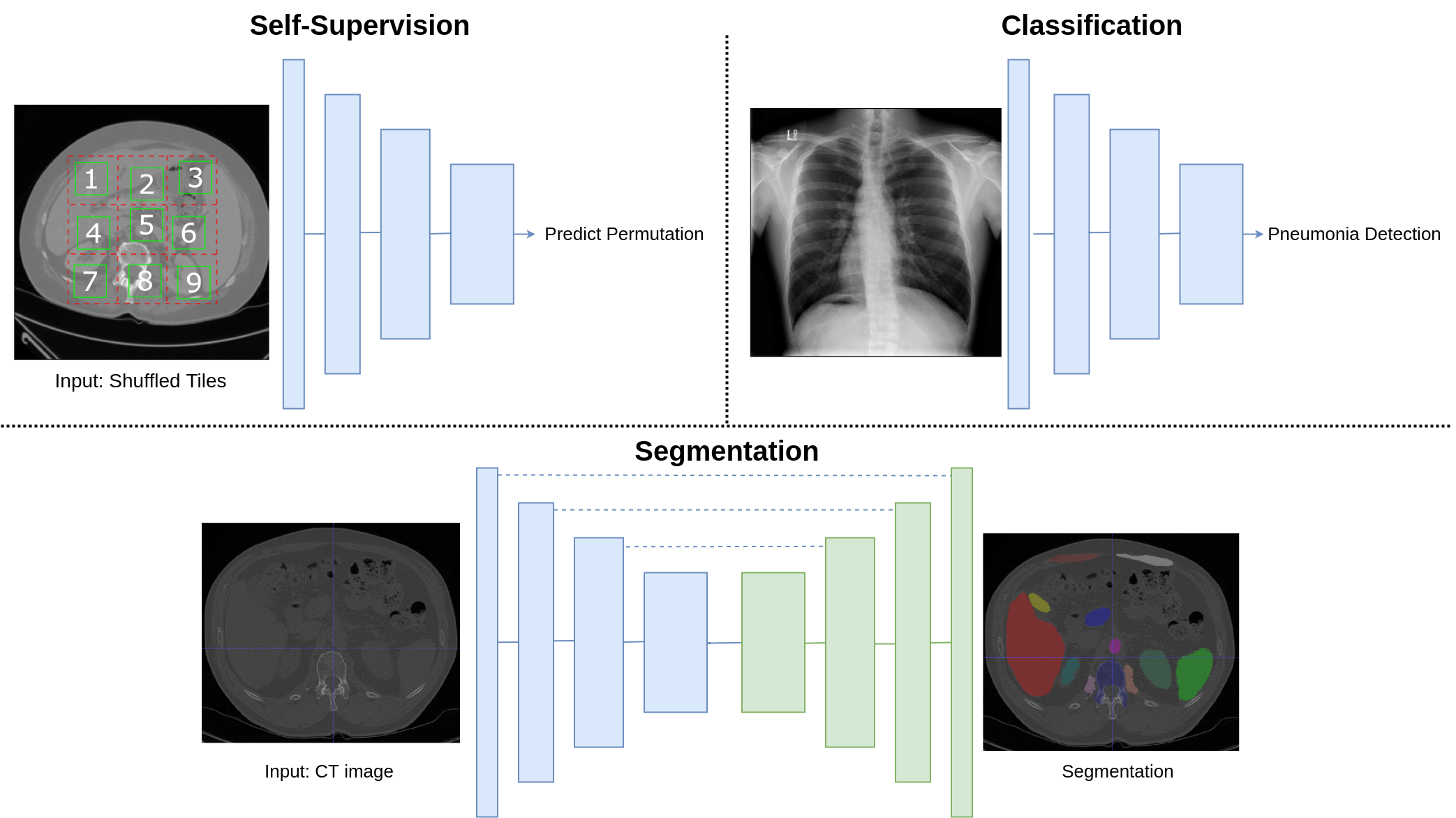}
    \caption{\footnotesize{Overview of self-supervised learning for JigsawNet. Learned feature representations are represented in blue. The learned feature representations can be directly transferred for the classification of pneumonia. For multi-organ segmentation, only the features in the encoder are transferred, the decoder depicted in green is initialized from scratch. Better viewed in color.}}
    \label{fig:models}
\end{figure}

\noindent\textbf{Self-Supervision:} In self-supervision, a network is trained to solve a pretext task that does not require accompanying labels. In this work, we evaluate the robustness and generalizability of JigsawNet \cite{norooziUnsupervisedLearningVisual2016} due to its superior performance in computer vision tasks. Furthermore, learning to solve jigsaw puzzles has been found to be effective to learn feature mapping of object parts as well as their spatial arrangement. This is an inherently appealing property in medical imaging for both classification and segmentation. For instance, learning the relative spatial arrangement of organs is necessary for organ segmentation. In JigsawNet, the learned features can be directly transferred to the target classification task. In contrast, for a segmentation task, features from the encoder are transferred and the decoder weights are initialized from scratch as shown in Fig. \ref{fig:models}.\\


\noindent\textbf{Full Supervision:} For fully-supervised classification networks, we use CBR-Small network architecture introduced in \cite{raghu2019transfusion}. This work demonstrated that deeper and complex networks like ResNet \cite{he2016deep} or Inception-V4 \cite{szegedy2017inception} are over-parametrized for medical imaging and that these networks achieve marginal performance compared to CBR-Small. For segmentation, a U-Net architecture is used \cite{ronneberger2015u}, given its performance in medical image segmentation.\\

\noindent\textbf{Implementation Details:} All models were trained using an Adam optimizer \cite{kingma2014adam}.  The learning rate is cosine annealed from $1 \times 10^{-3}$ to $1 \times 10^{-6}$. For classification, binary cross-entropy was used. For segmentation, a combination of dice loss and weighted cross-entropy was used in all experiments.


\section{Experiments and Discussion}


In this section, we describe the experiment settings, the data-sets, and the evaluation metrics used for our experiments. We first describe the comparative performance between the self-supervised and the fully-supervised networks on the clean data. Later, we show results in the presence of image corruptions to assess self-supervision robustness. Finally, we show experiments on the generalizability of both learning approaches.

\begin{table}[ht]
    \begin{minipage}{0.49\linewidth}
      
      \centering
        \begin{tabular}{lcc}
                & \multicolumn{2}{c}{\textbf{Metric}} \\ \hline
                & Accuracy         & AUROC            \\ \hline
Normal Training & 0.739           & 0.804          \\ \hline
Self-supervised          & \textbf{0.761}  & \textbf{0.821}  \\ \hline
\end{tabular}
\caption{\footnotesize{Average accuracy and AUROC for pneumonia classification.}}
\label{tab:clasres}
    \end{minipage}%
    \hfill
    \begin{minipage}{0.49\linewidth}
      \centering
        \begin{tabular}{lc}
                & \multicolumn{1}{c}{\textbf{Metric}} \\ \hline
                & \multicolumn{1}{c}{Dice Score}      \\ \hline
Normal Training & 0.900 $\pm$ 0.037                     \\ \hline
Self-supervised & \textbf{0.910 $\pm$ 0.029}            \\ \hline
\end{tabular}
\caption{\footnotesize{Average Dice and std for multi-organ segmentation.}}
\label{tab:segres}
    \end{minipage} 
\end{table}

\subsection{Data-sets}
We used two publicly available data-sets for classification and segmentation tasks respectively as described below.\\

\noindent\textbf{Pneumonia Detection Data-set:} For the classification task, we performed pneumonia detection on X-ray images from the Radiological Society of North America (RSNA) data-set \cite{wang2017hospital}. This data-set consists of 244,483 frontal X-rays with binary annotation for pneumonia obtained from radiologists and a test set of 3,024 images. To evaluate the performance of our networks under low data regimens, we use 1\% of the training data to train fully-supervised networks and fine-tune the target task after self-supervision. The remaining training data is used to train JigsawNet on the self-supervised task. We use the full test set for the evaluation performance.\\

\noindent\textbf{Multi-organ Segmentation Data-set:} The data-set used for multi-organ segmentation consists of CT scans from the VISCERAL data-set \cite{jimenez2016cloud}. The scans include CTs with and without contrast enhancement (ceCT), and two different fields of view: whole-body and thorax. Up to 20 different organs or structures are included in each scan. This data-set contains a Silver Corpus (SC) and a Gold Corpus (GC). The former contains annotations collected from different algorithms participating in the challenge. The latter describes annotations made by radiologists. We selected only those scans containing all 20 structures, resulting in 74 scans for SC and 22 for GC. Following a similar protocol for learning under data constraints, we use only 10\% of the SC (8 scans) to train the fully-supervised network and the target task after self-supervision. The self-supervision network trained to solve jigsaw puzzles used an additional 913 CT scans from the Grand Challenges \cite{challenges} data-set. This data-set includes CAD-PE \cite{cadpe}, EXACT \cite{lo2012extraction}, LiTS \cite{DBLP:journals/corr/abs-1901-04056}, LUNA \cite{setio2017validation}, Multi-atlas Labeling \cite{landman2015miccai}, CHAOS \cite{CHAOSdata2019} and SLIVER \cite{4781564}.




\subsection{Performance on Clean Data} 
We report the performance comparison between i) the network fined-tuned on the target task after learning the self-supervised task and ii) the network trained in a fully-supervised manner. Table \ref{tab:clasres} describes the average accuracy and the area under the receiver operating characteristic (AUROC) for the classification of pneumonia in the clean test data-set. We report a $p-$value $=0.0002$ and $p-$value $=0.003$ for accuracy and AUROC respectively using Wilcoxon signed-rank statistical test, which asserts that the performance improvements from the self-supervision are statistically significant. Likewise, Table \ref{tab:segres} shows the average Dice score along with its standard deviation for multi-organ segmentation. We obtain a $p-$value $=0.002$, confirming the statistical significance in the improvement provided by the self-supervision. We observe that self-supervision results in a consistent improvement compared to fully-supervised training. Nevertheless, these improvements are moderately low, considering the amount of data used for self-supervision. For instance, self-supervision yields a classification improvement of only 2\% over fully-supervised learning in accuracy and improvement of 1.6\% in the AUROC. For segmentation, self-supervised learning gives a nominal improvement of 1\% compared to fully-supervised learning. This raises questions concerning the utility of self-supervision given the similar performance of networks trained from scratch under a low number of training samples.

\begin{figure}[t]
  \centering
  \begin{minipage}[b]{0.49\textwidth}
    \includegraphics[width=\textwidth]{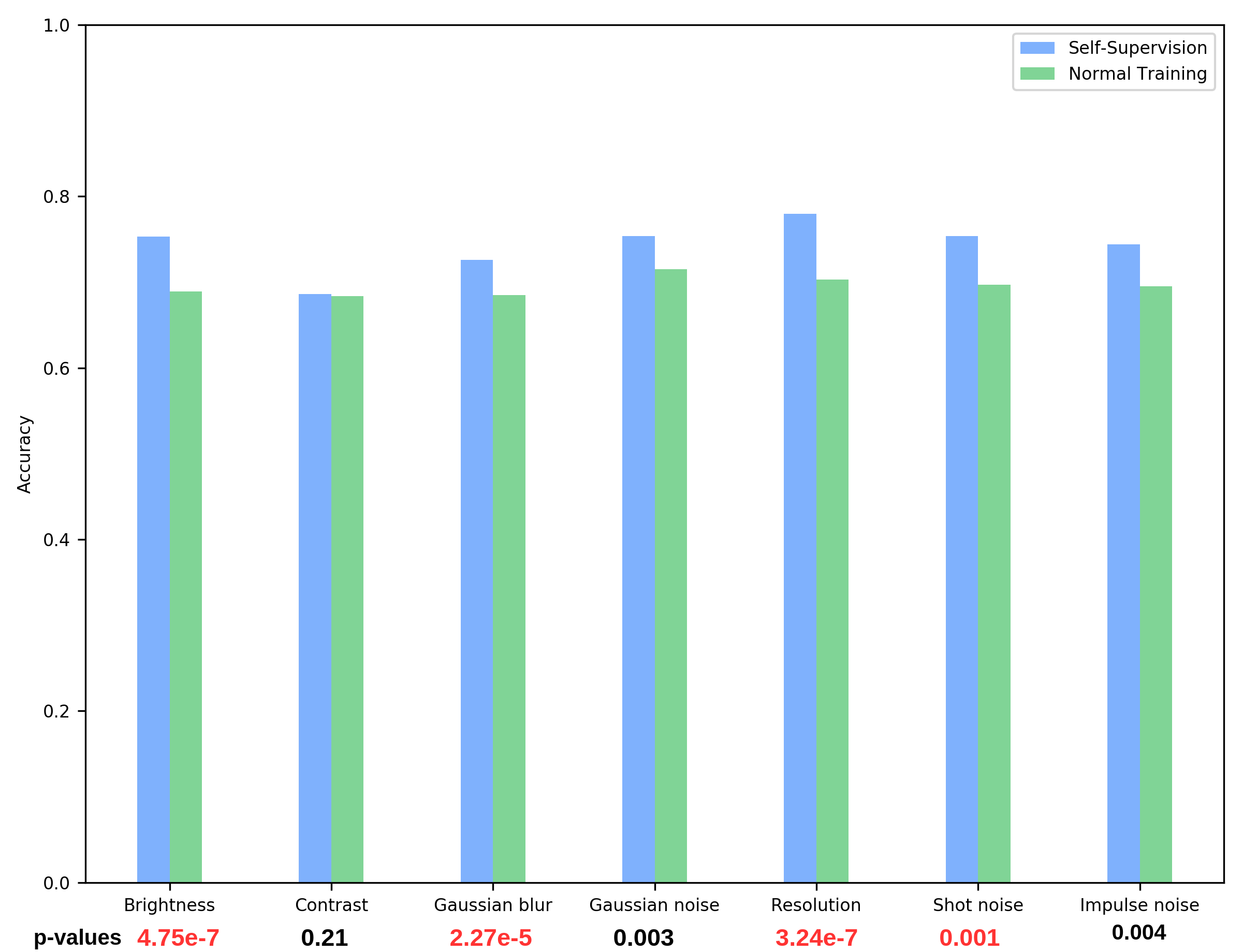}
    \caption{Performance comparison between normal training vs. self-supervision for classification. The horizontal axis describes the different image corruptions. The bars represent the average accuracy over five levels of severity for the given corruption.}
    \label{fig:accclass}
  \end{minipage}
  \hfill
  \begin{minipage}[b]{0.49\textwidth}
    \includegraphics[width=\textwidth]{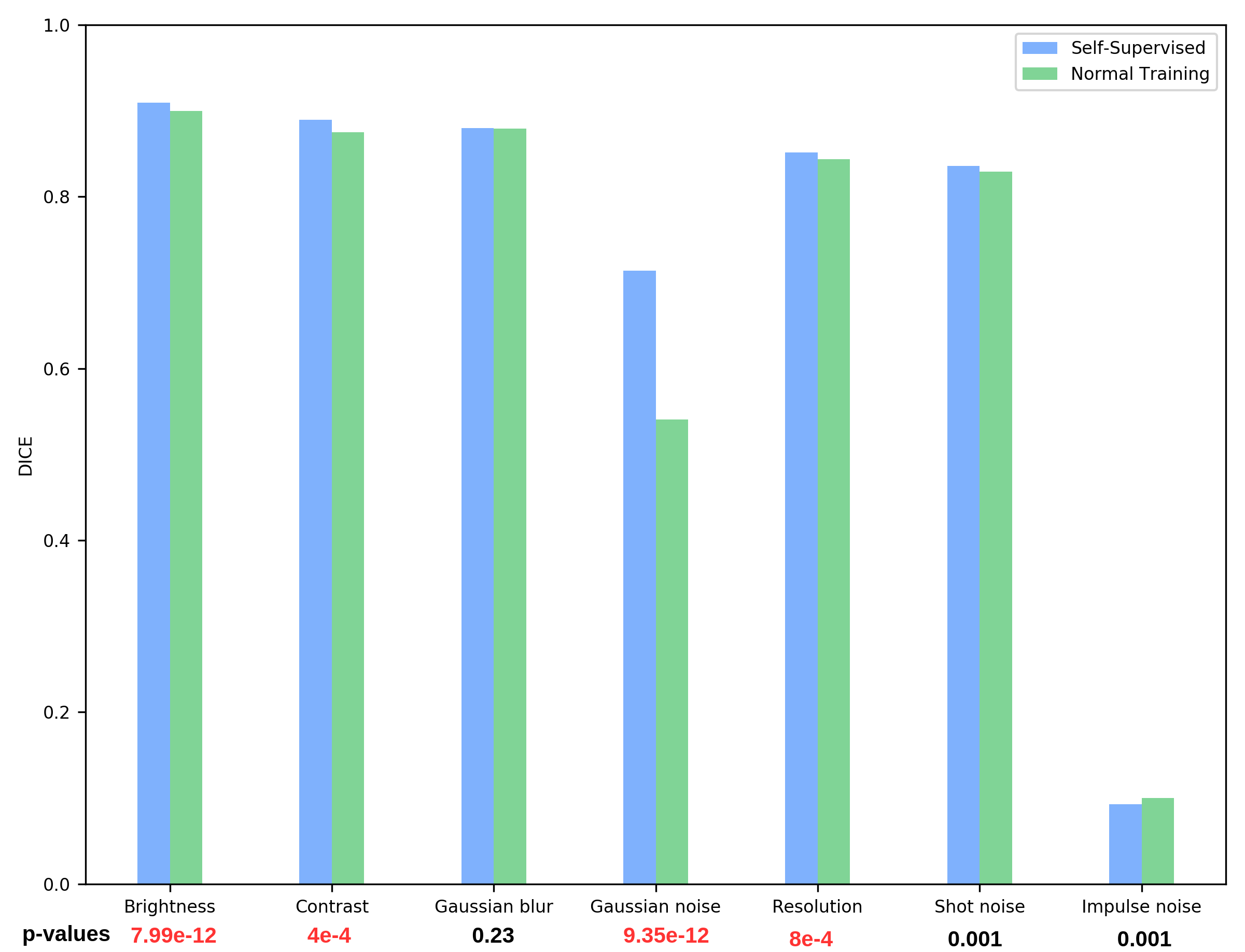}
    \caption{Performance comparison between normal training vs. self-supervision for segmentation. The horizontal axis describes the different image corruptions. The bars represent the average Dice over five levels of severity for the given corruption.}
    \label{fig:diceseg}
  \end{minipage}
\end{figure}

\subsection{Evaluating Robustness}

To address these questions, we evaluate networks trained via fully- and self-supervised learning in the presence of image corruptions. In our experiments, we evaluate a set of seven image corruptions described in the previous section and shown in Fig. \ref{fig:imagecorruption}. For each corruption, we apply five different levels of severity. These levels are found empirically for every image corruption and for every task (classification, segmentation) by keeping the lower bound performance higher than 0.49 using a validation set.\\

\noindent\textbf{Robustness in Classification:} For every corruption, we report the average accuracy over all images in the test set. Fig. \ref{fig:accclass} illustrates a comparison between CBR-Small architecture trained in a fully-supervised manner, referred to as  ``normal training", and the same network trained with self-supervision. We observe that in all image corruptions self-supervision is more robust, delivering higher average accuracy. In the same figure we report $p-$values $<0.001$ in red. Self-supervision shows high statistical significance in brightness, Gaussian blur, and resolution. In medical imaging, Gaussian blur can be related to a patient moving during acquisition and resolution is associated with data acquired at different imaging resolutions. Thus, the robustness results on these surrogate image corruptions make self-supervision an interesting approach.\\


\begin{figure}[t]
  \centering
    \includegraphics[width=0.8\textwidth]{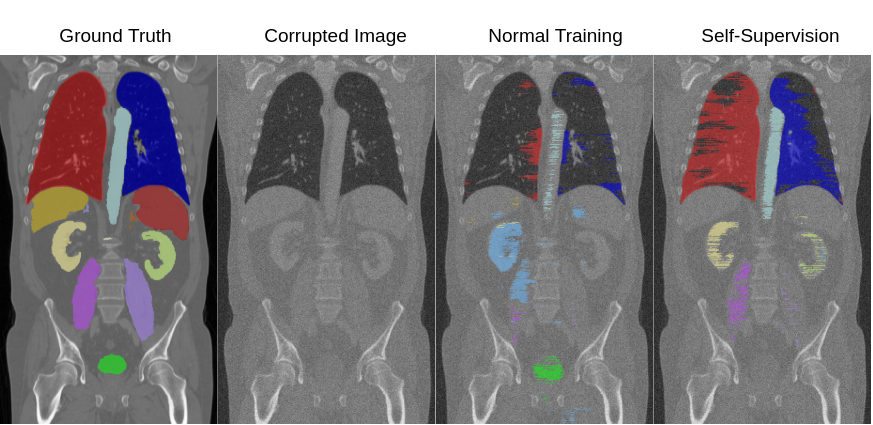}
    \caption{\footnotesize{Qualitative comparison for multi-organ segmentation in the presence of image corruption. From left to right: the original image with the ground truth segmentation, the corrupted image, the predictions from normal training, and the predictions from self-supervision.}} 
    \label{fig:segcompare}
\end{figure}


\noindent\textbf{Robustness in Segmentation:} We also perform the image corruption experiments for the multi-organ segmentation task using UNet architecture. Fig. \ref{fig:diceseg} describes the overall Dice score over corruption for normal training and self-supervision. $P-$values $<0.001$ are reported in red, showing that brightness, contrast, Gaussian noise, and resolution yielded a statistically significant result. Fig. \ref{fig:diceseg} shows that self-supervision has superior segmentation performance in the presence of data corruption. Impulse noise is the only corruption for which self-supervision has lower performance. Nevertheless, both networks are extremely sensitive to this type of noise, dropping Dice scores to $\sim$ 0.1. A qualitative comparison between self- and fully-supervised learning is shown in Fig. \ref{fig:segcompare} when we apply Gaussian noise to the input image, which can arise during image acquisition as the result of a trade-off between radiation dose and signal quality. We observe that the predictions of normal training fail in segmenting most of the organs. Notice that self-supervision does not flip labels between different structures, even in the presence of data corruption.

Regarding the robustness of self-supervision against image corruption, our experiments confirm that features learned with a self-supervised approach are more robust to data corruption. We owe this to the fact that self-supervised networks see a larger number of images from different data-sets during self-supervision compared to fully-supervised learning. Furthermore, we attribute its superior performance to the fact that in JigsawNet the network is forced to learn spatial arrangement, resulting in stronger and more robust features.



\begin{table}[t]
	\begin{minipage}{0.6\linewidth}
	\centering
    \includegraphics[width=0.8\textwidth]{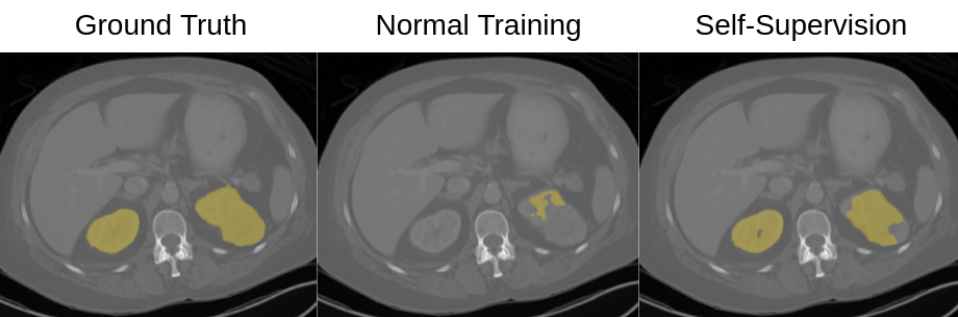}
    \caption{\footnotesize{Qualitative results in network generalizability. From left to right: ground truth, normal training prediction, self-supervision prediction.}}
    \label{fig:segkits}
	\end{minipage}\hfill
	\begin{minipage}{0.38\linewidth}
    \centering
		\begin{tabular}{ll}
                & \multicolumn{1}{c}{\textbf{Metric}} \\ \hline
                & \multicolumn{1}{c}{Dice Score}      \\ \hline
Normal Training & 0.486 $\pm$ 0.246                            \\ \hline
Self-supervised           & \textbf{0.670} $\pm$ 0.209                      \\ \hline
\end{tabular}
\caption{\footnotesize{Average segmentation performance from deploying the trained networks on KITS2019. $p-$value $=0.0002$}.}
\label{tab:kidneys}
	\end{minipage}
\end{table}

\subsection{Evaluating Generalizability} 
To evaluate the generalizability of self-supervision, we benchmark the networks originally trained for multi-organ segmentation on the KITS2019 data-set \cite{heller2019kits19}. We are interested in the performance of both normal training and self-supervision when they are deployed in a different domain. Table \ref{tab:kidneys} shows the average Dice score and standard deviation for kidney segmentation. Normal training gives an average Dice of 0.486 while self-supervision increases the average Dice by 18.4\%, resulting in a Dice score of 0.670. In Fig. \ref{fig:segkits} we show a qualitative evaluation. Self-supervision segments both kidneys, while normal training totally fails in one of the kidneys. These findings confirm our hypothesis on the generalizability of self-supervision. Learning a pretext task and using a large number of images during training results in features that are useful for domain adaptation.


\section{Conclusions}

In this work, we have demonstrated that despite the marginal differences in performance between fully-supervised learning and self-supervision, the  benefits of self-supervision are revealed when networks are bench-marked on imperfect data. We have demonstrated this in the context of medical imaging for two tasks, pneumonia detection and multi-organ segmentation. Moreover, we have shown the improved robustness and generalizability of self-supervision across domains compared to fully-supervised learning. Finally, this work suggests that robust feature representations can potentially be improved with self-supervision and opens a new direction towards finding novel self-supervision paradigms for robustness and generalizability rather than developing self-supervision approaches as a mechanism of catching up to fully-supervised performance.\\

\bibliographystyle{splncs}
\bibliography{chrisbib}
\end{document}